\documentclass[conference]{IEEEtran}
\IEEEoverridecommandlockouts
\pdfoutput=1
\usepackage{cite}
\usepackage{amsmath,amssymb,amsfonts}
\usepackage{algorithmic}
\usepackage{graphicx}
\usepackage{textcomp}
\usepackage{xcolor}
\usepackage{tikz}
\usepackage{balance}
\usepackage[inline]{enumitem}

\usepackage{pgfplots}
\usepackage{pgfplotstable}
\pgfplotsset{compat=1.7}
\usepgfplotslibrary{groupplots}
\usepackage{soul} 
\sethlcolor{orange}
\usepackage[normalem]{ulem} 

\usetikzlibrary{external}

\usepackage{amsmath}
\def\BibTeX{{\rm B\kern-.05em{\sc i\kern-.025em b}\kern-.08em
    T\kern-.1667em\lower.7ex\hbox{E}\kern-.125emX}}

\usepackage{booktabs}
\usepackage{siunitx}

\begin{document}

\title{Machine Learning Methods for Device Identification Using Wireless Fingerprinting\\
\thanks{The work is supported in part by European Commission’s Horizon 2020
Research and Innovation Programme Grants No. 871518 and 856967.}
}

\author{\IEEEauthorblockN{Srdjan Sobot, Vukan Ninkovic, Dejan Vukobratovic}
\IEEEauthorblockA{\textit{Department of Power, Electronics and Communications} \\
\textit{Faculty of Technical Sciences, University of Novi Sad}\\
Novi Sad, Serbia \\
\{srdjansobot, ninkovic, dejanv\}@uns.ac.rs}
\and
\IEEEauthorblockN{Milan Pavlovic, Milos Radovanovic}
\IEEEauthorblockA{\textit{Department of Mathematics and Informatics} \\
\textit{Faculty of Sciences, University of Novi Sad}\\
Novi Sad, Serbia \\
\{milan.pavlovic, radacha\}@dmi.uns.ac.rs}
}

\maketitle

\begin{abstract}
Industrial Internet of Things (IoT) systems increasingly rely on wireless communication standards. In a common industrial scenario, indoor wireless IoT devices communicate with access points to deliver data collected from industrial sensors, robots and factory machines. Due to static or quasi-static locations of IoT devices and access points, historical observations of IoT device channel conditions provide a possibility to precisely identify the device without observing its traditional identifiers (e.g., MAC or IP address). Such device identification methods based on wireless fingerprinting gained increased attention lately as an additional cyber-security mechanism for critical IoT infrastructures. In this paper, we perform a systematic study of a large class of machine learning algorithms for device identification using wireless fingerprints for the most popular cellular and Wi-Fi IoT technologies. We design, implement, deploy, collect relevant data sets, train and test a multitude of machine learning algorithms, as a part of the complete end-to-end solution design for device identification via wireless fingerprinting. The proposed solution is currently being deployed in a real-world industrial IoT environment as part of H2020 project COLLABS. 

\end{abstract}


\section{Introduction}

Explosion in the number of connected IoT devices brought increasing concerns for IoT systems security \cite{Jing_2014}. Industrial IoT systems are particularly vulnerable, since IoT devices may participate in mission-critical industrial control processes \cite{Gebremichael_2021}. In recent years, the IoT security methods that exploit a combination of specific features unique to the device, called fingerprints, and machine learning (ML) algorithms, became increasingly popular \cite{Soltanieh_2020, Xu_2016, Polak_2011, Jana_2010, Wang_2017, Neumann_2012, Gao_2010, Msadek_2019}. In particular, wireless fingerprints (WF) may be extracted at the physical (PHY) layer of a wireless receiver, based on the signals received from different wireless IoT devices \cite{Soltanieh_2020, Xu_2016}. 

The most informative WF signals are raw received discrete-time complex baseband waveforms. These signals reveal the radio-frequency (RF) hardware signatures imprinted by individual IoT devices \cite{Polak_2011}. However, capturing such signals is costly in terms of the data volume, while access to raw signals is severely limited in commercial radio interfaces. Instead, the signals such as channel state information (CSI) or radio channel strength statistics are commonly captured (see Sec. II) \cite{Wang_2017}. Such signals are accessible in commercial devices and their data volume is manageable. However, the extracted features depend on the device location and the surrounding environment, instead of unique RF hardware signatures.

In this paper, we present our end-to-end design of the WF-based device identification system. The system architecture comprises a WF module that captures and extracts relevant WFs, and a ML module that applies ML algorithms (MLAs) to perform device identification. In contrast to previous works, we take a more comprehensive design and evaluation approach, where we consider the design of WF modules for the most popular wireless IoT technologies, and we develop ML modules employing a large collection of popular MLAs.

We develop and integrate WF modules based on the 3GPP Narrowband-IoT (NB-IoT) cellular IoT and IEEE 802.11n Wi-Fi standards suitable for industrial IoT deployment as part of private 4G LTE or Wi-Fi networks. The WF modules are implemented on custom-designed NB-IoT and Wi-Fi devices and designed to extract various WF information. The WF modules stream extracted WF data from IoT devices to a remote cloud-based ML module using commonly used RESTful framework.

Comprehensive ML module based on both classical and deep learning methods is implemented as a cloud-based service. The ML module employs representative methods for supervised device identification due to the availability of device IDs in the collected WF data sets. Due to different WFs collected from NB-IoT and Wi-Fi devices, different MLAs are considered for the two scenarios. The system is integrated and tested in our labs and an extensive device identification performance is presented and visualised. The setup is currently being migrated to a real-world industrial IoT scenario, where it will be integrated as part of the comprehensive IoT cyber-security platform as part of the H2020 COLLABS project \cite{collabs}.


\section{Background and System Model}

\subsection{Wireless Fingerprinting Techniques}

Depending on the protocol stack layer where fingerprints are captured, we distinguish WF based on:        

\subsubsection{Radio Frequency (RF) Fingerprinting} 
\label{RF_finger} The features obtained from ``raw'' signals, i.e., RF waveforms at the receiver PHY layer, are suitable for device identification tasks \cite{Soltanieh_2020, Xu_2016, Wang_2017}. In certain scenarios, the extracted signatures can be also used for device localization. WFs obtained from the receiver PHY layer are commonly classified into two groups\cite{Xu_2016}: \begin{enumerate*}
\item location-dependent, and\label{item:LD}
\item location-independent features. \label{item:LI}
\end{enumerate*}

\begin{figure}[t]
	\centering
	\includegraphics[width=0.95\linewidth]{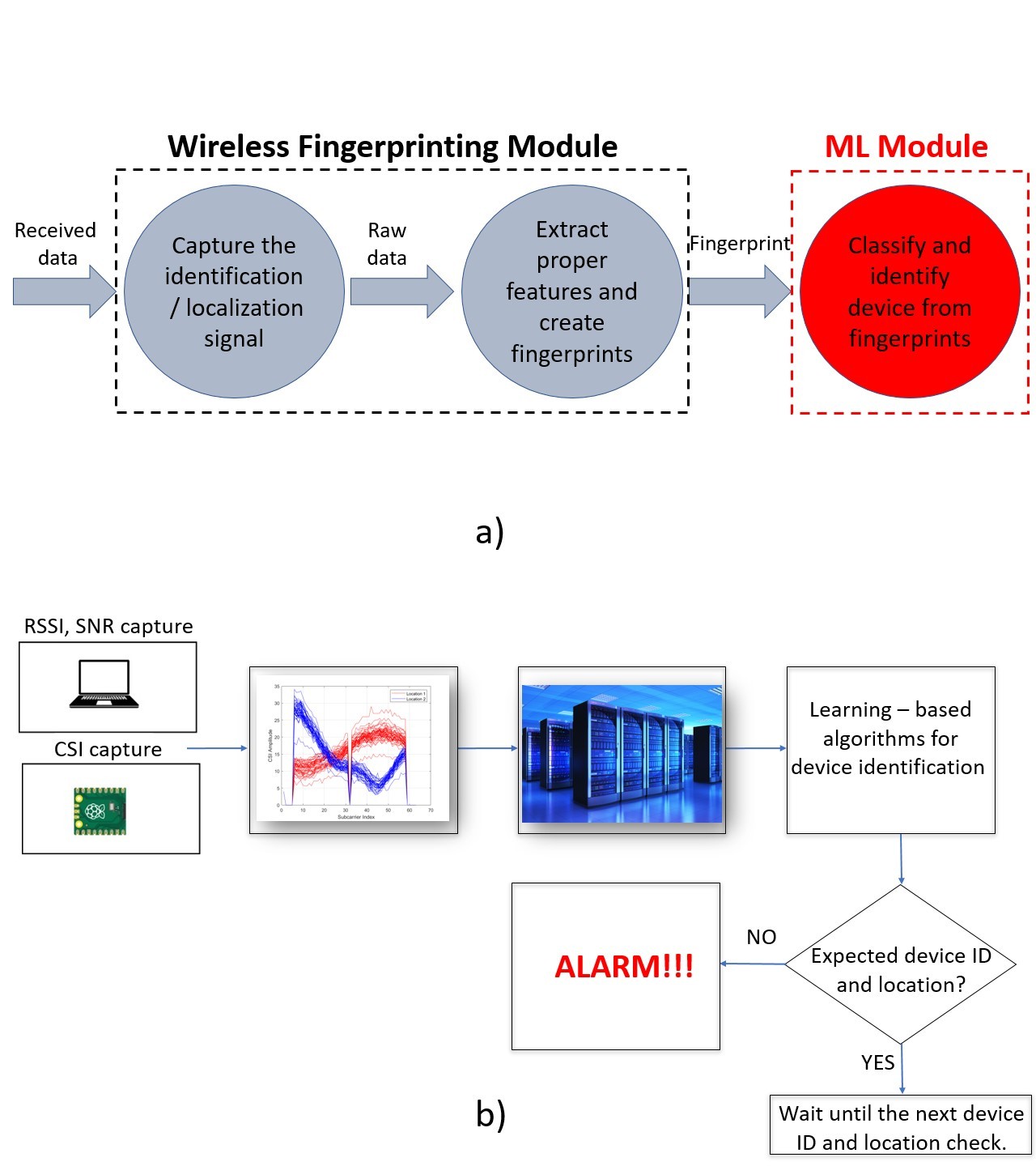}
	\caption{PHY layer identification system: a) Generic system architecture, b) Detailed flowchart of the presented implementation.}
	\label{Fig_SA}
\end{figure}

Received Signal Strength Indicators (RSSI), Reference Signal Received Power (RSRP), Reference Signal Received Quality (RSRQ) and Signal to Noise Ratio (SNR) and CSI are common examples of location-dependent features. RSSI, RSRP, RSRQ and SNR depend on the transmission power and channel (environment) conditions and are represented as a real number (e.g., a value in dB) averaged per data packet \cite{Xu_2016}. RSSI, RSRP, RSRQ and SNR provide promising performance in device identification and localization tasks. However, in the situations where devices are close to each other, their values can be less discriminative and, in multipath environments, unstable \cite{Wang_2017}. In other words, for applications demanding high performance and robustness, finer-grained features that exploit OFDM signal structure are needed. This is possible by extracting CSI estimates that represent individual channel gains estimated per each OFDM subcarrier.

Location independent features are usually based on hardware imperfections, which arise during the production process of individual devices \cite{Xu_2016}. These imperfections, although small, can provide unique device signatures. For example, the power amplifier represents the final element of the RF chain at the transmitter, and as such, their signatures are hard to be modified by attackers \cite{Xu_2016}. As a consequence, imperfections of such components can be exploited as part of the device identification \cite{Polak_2011}. However, accessing signals from which it is possible to extract such signatures is usually hard. Ideally, one would need to have access to samples of discrete-time complex baseband samples (I/Q samples) at the output of the receiver RF part, which is typically not the case.      
\subsubsection{Traffic Fingerprinting} The unique features extracted from the data link/MAC layer and/or higher protocol layers can be used as a basis for the \textit{traffic fingerprinting}. According to \cite{Xu_2016}, MAC layer features are easy to extract and the MAC layer implementation is frequently vendor specific (i.e., some details are underspecified by the standard and depend on the vendor implementation). This characteristic is explored in \cite{Jana_2010}, where clock skew of access point (AP) is used as a device feature, and calculated through the IEEE 802.11 Time Synchronization Function (TSF) timestamps sent in the beacon/probe response frames. Also, the effectiveness of different MAC layer features in tasks of interest, such as transmission time and rate, frame size and medium access time, are evaluated in \cite{Neumann_2012}.

Unique fingerprints can be obtained from higher layers that provide different information about traffic characteristics. In \cite{Gao_2010}, inter-arrival time of TCP/UDP packets is used as a signature in order to distinguish different APs. Also, through the most used destination ports for TCP/UDP packets, vulnerability of the IoT devices is explored and analyzed in \cite{Msadek_2019}.   

\subsection{System Architecture}

The device fingerprinting approach presented in this paper is based on PHY layer features described in Section \ref{RF_finger}. We follow the system architecture that performs three tasks \cite{Soltanieh_2020}: \begin{enumerate*}
\item Capture the identification signal, 
\item Extract relevant features and create fingerprints, and
\item  Classify and identify a device from its fingerprints.
\end{enumerate*}  
The overall architecture is shown in Fig. \ref{Fig_SA} a), where the process of data collection and extraction of fingerprints is merged into the \textit{Wireless Fingerprinting (WF) Module}, while the algorithmic part for device identification is separated into the \textit{Machine Learning (ML) Module}. The ML module can be executed as a remote edge or cloud application. 

In the presented setup, the WF module represents a data generation mechanism attached directly to the IoT devices. Its task is to collect, extract, format and stream the fingerprinting information to be further processed by the ML module. The task of the ML module is to analyze the behaviour of IoT devices using appropriately selected ML algorithms based on the input WF features. The main problems that the ML component tries to solve by using WF features are device identification and localization, as well as detection of unknown devices. Depending on the available data, MLA can be trained in a supervised, semi-supervised, or unsupervised fashion. The focus of this paper is on supervised device identification. The fine-grained representation of the proposed approach is shown, in flowchart form, in Fig. \ref{Fig_SA} b). 
\section{Wireless Fingerprinting Module}

The system architecture presented above is generic and may include different realisations of WF and ML modules. In order to explore its versatility to different communication standards, scenarios and WF fingerprints, the architecture is implemented and tested on two popular IoT communication paradigms: \begin{enumerate*}
\item Cellular IoT communication system (3GPP NB-IoT) and
\item  Wi-Fi IoT communication system (IEEE 802.11n).
\end{enumerate*} 

\subsection{3GPP NB-IoT}
\label{NB_IoT}
NB-IoT standard is based on a multi-carrier OFDM signal. In the frequency domain, it consists of 12 OFDM subcarriers separated by a 15 kHz subcarrier spacing. In the time domain, the signal consists of consecutive frames of 10~ms duration, divided into 1~ms subframes, which are further divided into 14 OFDM symbols. A single OFDM carrier during the period of a single OFDM symbol represents a resource element that carries a single QPSK or BPSK modulated symbol. Selected resource elements carry reference signals which are used to measure radio-specific information.
\begin{figure}[t]
	\centering
	\includegraphics[width=0.8\linewidth]{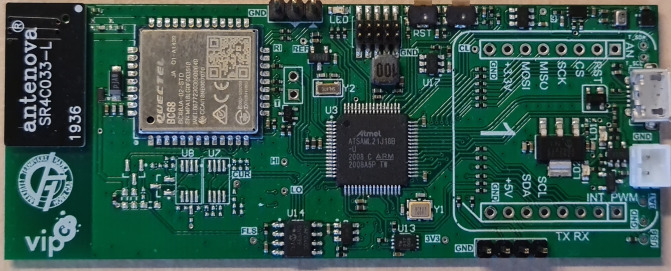}
	\caption{Custom-designed NB-IoT module for WF collection.}
	\label{Fig_nbiot}
\end{figure}
For NB-IoT scenario, we use the following location-dependent fingerprints to identify devices: RSSI, RSRP, RSRQ and SNR \cite{3gpp}. The radio parameters under consideration are averaged over the data packet transmission. 

\subsubsection{Data Set Generation}
\begin{figure}[t]
	\centering
	\includegraphics[width=1\linewidth]{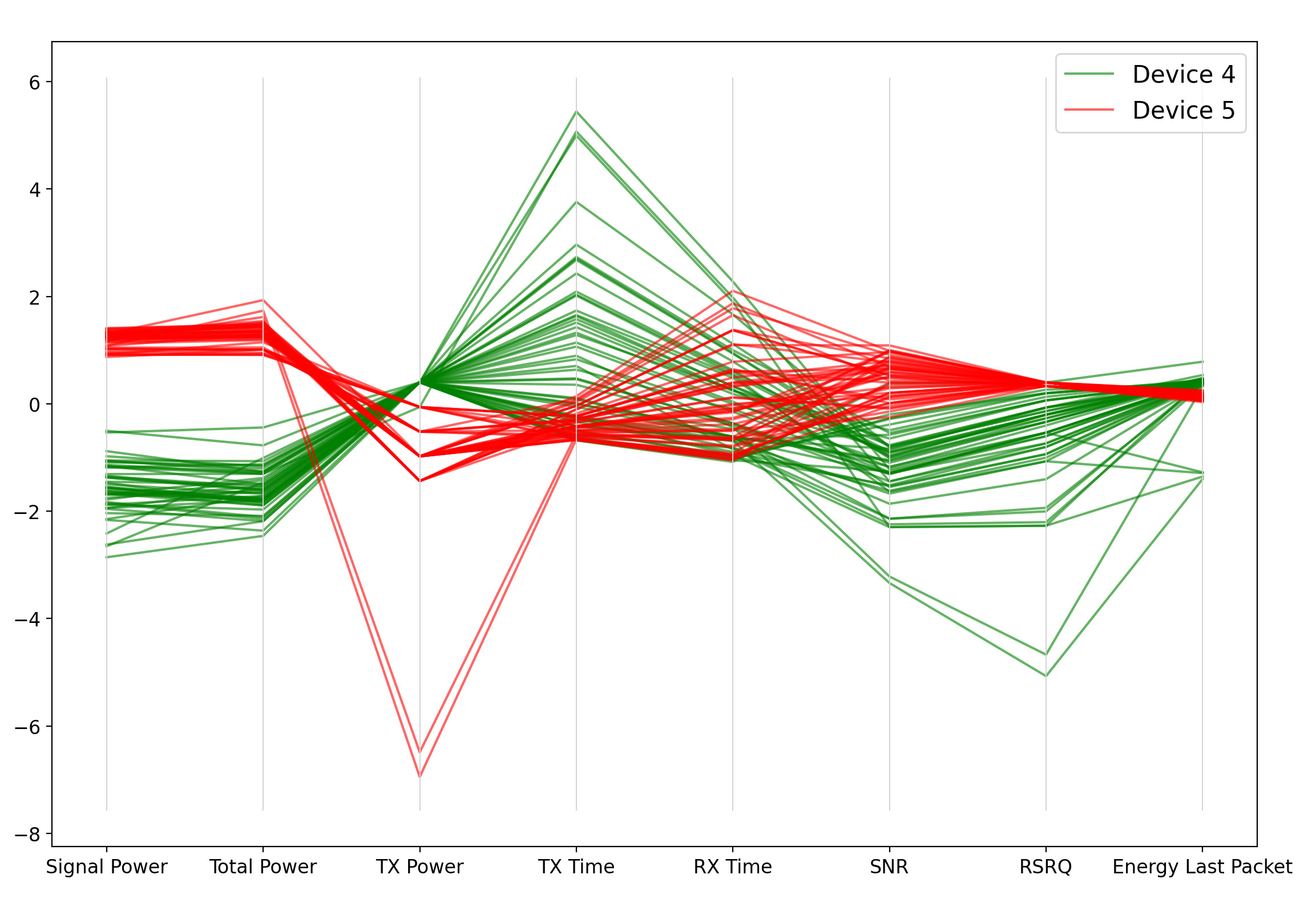}
	\caption{Parallel coordinates chart of standardized features from 3GPP NB-IoT data set. Each vertical line represents a corresponding attribute, while each wireless fingerprint is plotted horizontally across all attributes. Devices 4 and 5 are presented in green and red, respectively, with 50 fingerprints each.}
	\label{Fig_NB_Signals}
\end{figure}

\begin{figure}[t]
	\centering
	\includegraphics[width=0.9\linewidth]{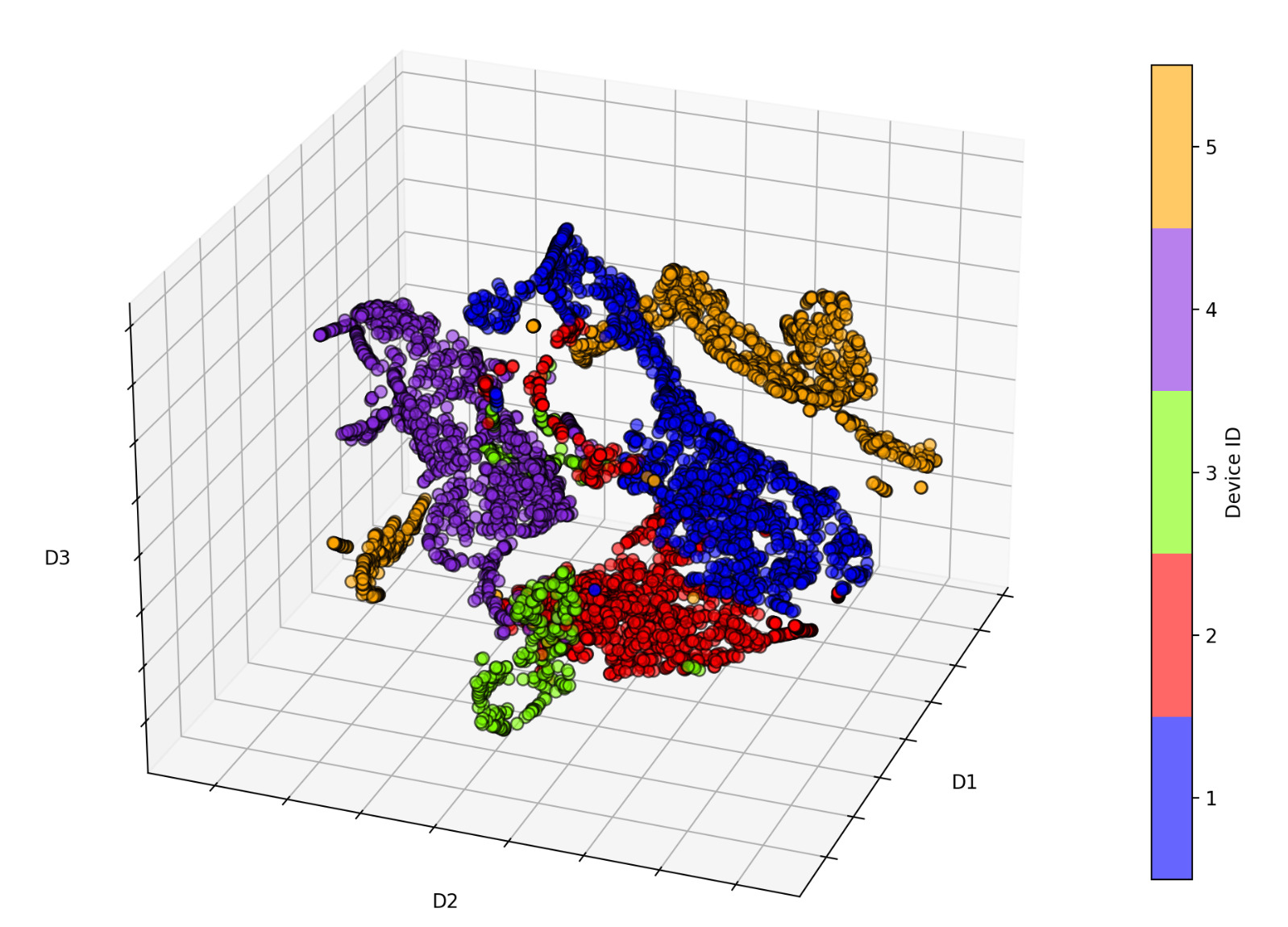}
	\caption{Visualization of 3D feature space on 3GPP NB-IoT data set. One point corresponds to one wireless fingerprint, while the color indicates belonging to the corresponding device.}
	\label{Fig_NB_TSNE}
\end{figure}

To generate the dataset, we used custom-designed NB-IoT modules shown in Fig. \ref{Fig_nbiot}. Five devices were placed in an indoor lab, about 500 m away from the macro-cellular base station. The dataset is collected using UE monitoring software that collects radio measurements averaged across each of 5425 data packets. 3D space of the features, created with the t-SNE dimensionality reduction technique \cite{TSNE}, is presented in Fig. \ref{Fig_NB_TSNE} for each device. The most informative radio parameter in our measurements is RSRP, as shown in Fig.  \ref{Fig_NB_Signals} (RSRP parameter is marked as Signal Power) for two selected NB-IoT devices. 

\subsection{IEEE 802.11n}

In the Wi-Fi domain, we consider the most common IEEE 802.11n standard \cite{Perahia_2008}, typically used in industrial environments \cite{Tramarin_2016}. Similarly to NB-IoT, IEEE 802.11n is an OFDM-based standard. In industrial Wi-Fi environment, wireless devices are usually closer to each other than in cellular networks, thus finer-grained data is needed. In this work, a low-level channel state information (CSI) is extracted from the PHY layer. CSI values contain complex-valued average channel gains estimated per each OFDM subcarrier. In IEEE 802.11n standard, each OFDM symbol has $N=64$ subcarriers, out of which 52 is reserved for the data \cite{Tramarin_2016}, while the rest represent the guard band.

\subsubsection{Data Set Generation}

\begin{figure}[t]
	\centering
	\includegraphics[width=0.9\linewidth, angle=90]{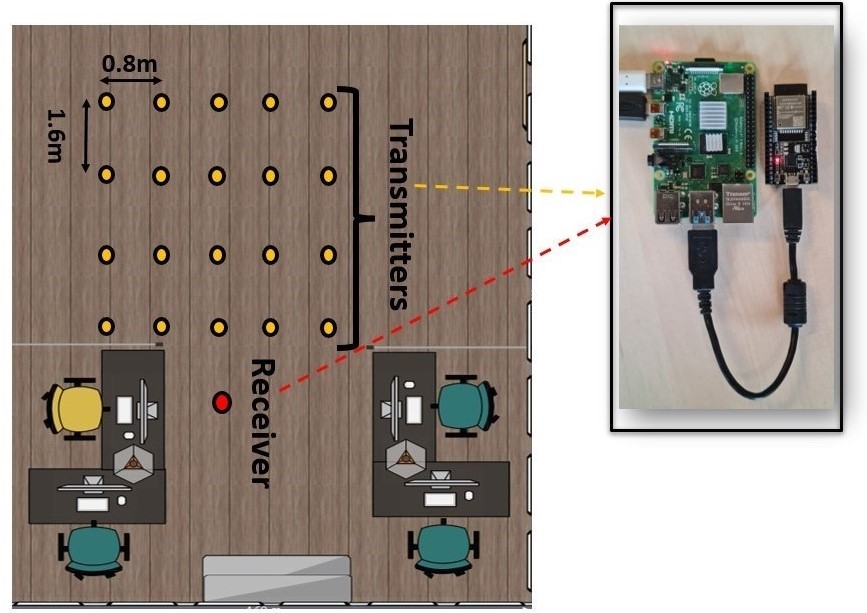}
	\caption{IoT devices (Tx) and AP (Rx) positions in the predefined grid. Each Tx/Rx is an ESP32+RPi device.}
	\label{Fig_Grid}
\end{figure}

In order to collect data set, we deploy our setup in an indoor environment, placing the IoT devices on 20 predefined grid points, as shown on Fig. \ref{Fig_Grid}. The IoT devices and the AP are stationary. Both IoT devices and the AP use ESP32 platforms with integrated IEEE 802.11n chipset for data generation and collection. The devices are powered by their host, a RaspberryPi (RPi)-based industrial IoT node, as shown in Fig. \ref{Fig_Grid}.

\begin{figure}[t]
	\centering
	\includegraphics[width=0.9\linewidth]{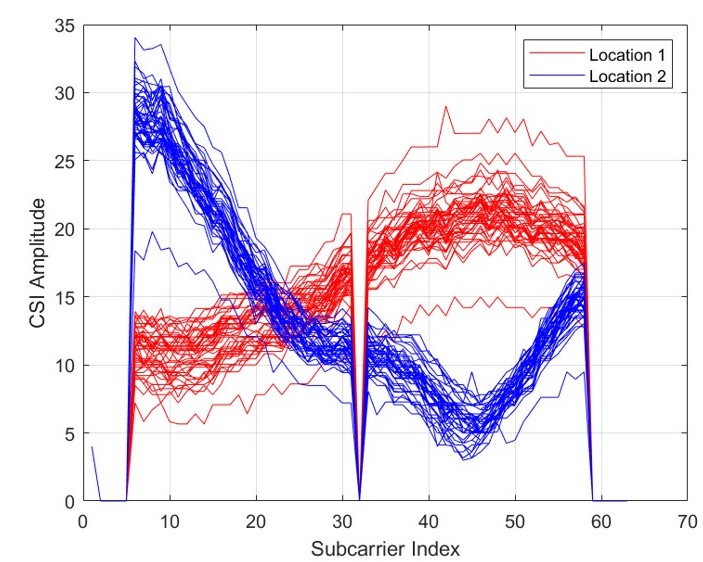}
	\caption{Received CSI amplitudes from 2 different locations. One curve represents one packet, from each location 50 packets are received.}
	\label{Fig_AMP}
\end{figure}

\begin{figure}[t]
	\centering
	\includegraphics[width=0.9\linewidth]{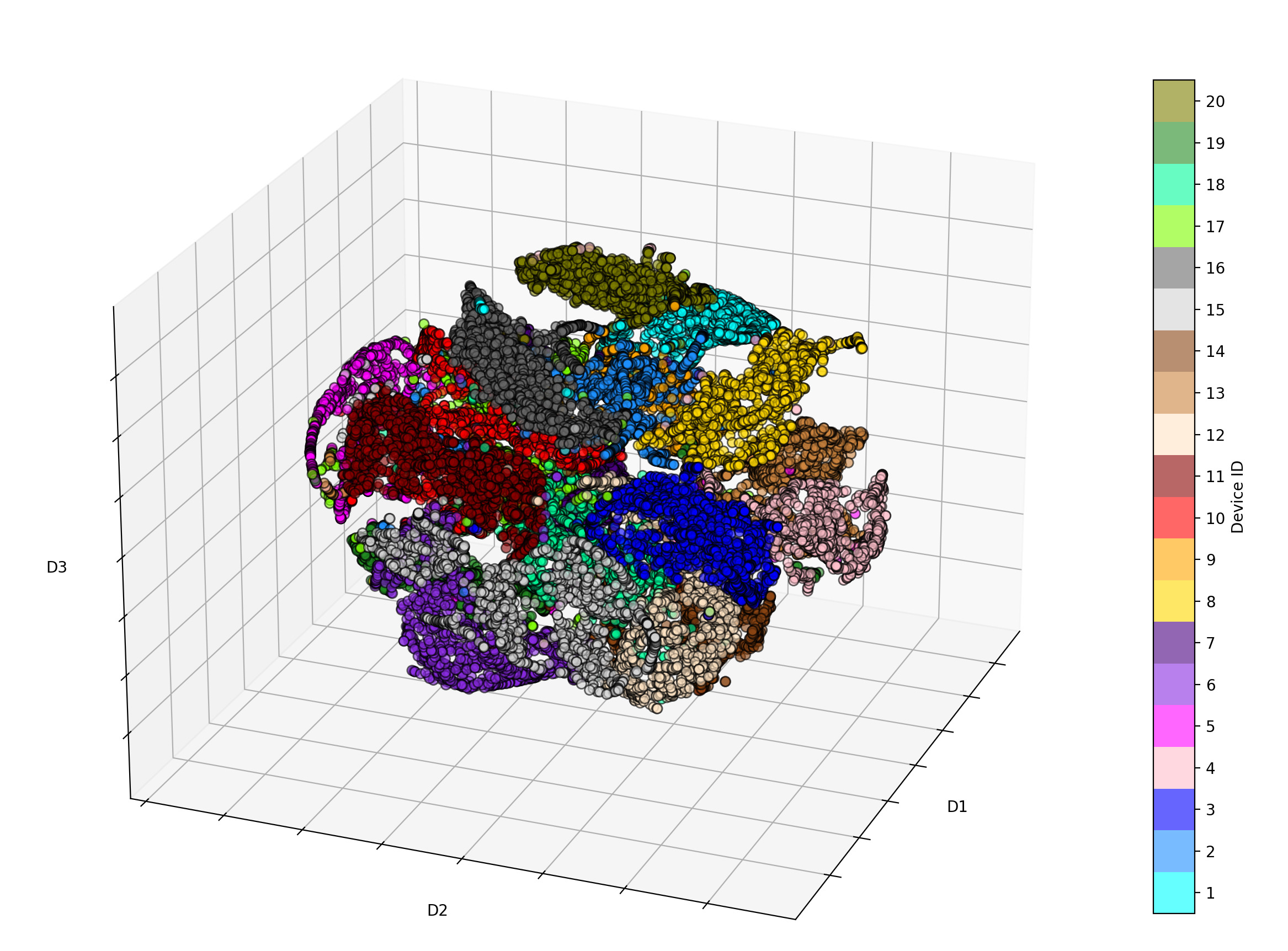}
	\caption{Visualization of 3D feature space on IEEE 802.11n data set. Each point represents one wireless fingerprint, while the color indicates belonging to the corresponding device.}
	\label{Fig_ESP_TSNE}
\end{figure}

During the offline phase, the average CSI values of about 1250 packets are collected per ESP32 device, using \textit{ESP32 toolkit} \cite{CSI_2020} to extract amplitudes of the received CSIs. An example of such data is presented in Fig. \ref{Fig_AMP} for 2 different IoT devices. The overall data set consists of 24722 ($|CSI|$, \textit{id}, \textit{location}) triplets, where \textit{id} and \textit{location} represent labels for identification and possible localization tasks. 3D feature space for 20 different devices used in this experiment is illustrated in Fig. \ref{Fig_ESP_TSNE}.

In the online phase, transmitters simultaneously send packets to the receiver, which forwards their amplitudes in bursts towards the server, where parameters of the trained ML algorithm (in offline phase) are deployed. 

\section{Machine Learning Module}

The ML Module is organized around a framework consisting of classical machine learning and deep learning (DL) models that enable the analysis of the behavior of multiple IoT devices based on WF. The three main challenges ML module addresses are device identification, localization, and detection of unknown devices.

The \textit{device identification} task represents the recognition of a device based on its WFs. In a supervised setup, this problem can be transformed to a multi-class classification problem, where one class belongs to WFs from a single device. On the other hand, unsupervised approaches tends to cluster together similar WFs from multiple logical devices (based on their identifiers) and map them as the same physical device \cite{Xu_2016}.

The \textit{device localization} task aims to determine the location of the device based on its WFs. From the ML perspective, the prediction of coordinates can be presented as a regression problem in the case when coordinates are available.

The task of \textit{detecting unknown devices} is to distinguish between known and unknown devices. Anomaly detection algorithms makes possible to create a frontier delimiting the contour of WFs distribution from known devices. Everything outside the region is consider to be unknown device.

The main focus of this paper is supervised device identification due to the availability of device IDs in the aforementioned data sets. There are various ways how WF representations can be built and thus improve the performance of the ML model. A single WF can be presented in the form of a univariate/multivariate time series or as a set of attributes without time dependency.

We created input WF vectors without exploiting time dependency from the data set in Section III-A (3GPP NB-IoT). Therefore, the following non-sequential MLAs \cite{MLA} were applied: Naive Bayes (NB), where the likelihood of the features is assumed to be Gaussian; Instance-based K-Nearest Neighbors (KNN) classifier; Linear algorithm Logistic Regression (LR); Support Vector Machine (SVM) classifier; Interpretable Decision Tree (DT); Tree-based ensemble Random Forest (RF); Highly-efficient Gradient Boosting Machines (GBM) \cite{GBM}.

In the scenario from Section III-B (IEEE 802.11n), WF representations are univariate time-series, and thus we performed the following sequence-based MLAs: Word Extraction for Time Series Classification (WEASEL) \cite{WEASEL} for feature extraction and RF as a classifier; Random Convolutional Kernel Transformation (ROCKET) \cite{ROCKET} for transforming time series into feature vectors, followed by RF; Plain Elman Recurrent Neural Network (RNN) \cite{RNN}; Recurrent architecture based on Long Short-Term Memory (LSTM) \cite{LSTM} with an additional global pooling layer that aggregates all outputs of the network and passes them to the fully-connected layer; Aggregated Residual Network (RESNEXT) \cite{RESNEXT} with the 1D convolutions; Selective Kernel Network (SKNET) \cite{SKNET}, a convolutional architecture with a dynamic selection mechanism founded on attention. 

\section{Performance Evaluation}

Figs. \ref{Fig_NB_TSNE} and \ref{Fig_ESP_TSNE} show that the fingerprints of each device form dense clusters. Therefore, WFs of one device tend to be similar to each other and different from WFs of other devices. In addition, we spotted that fingerprints from device 4 have a value interval that is significantly different from the corresponding interval of device 5 for attribute Power Signal (RSRP) in Fig. \ref{Fig_NB_Signals}. Hence, this indicates that the signal power (RSRP) is a strong predictor.

Within this work, we performed two experiments. The first experiment is the application of non-sequential MLAs from Section IV to the data set described in Section III-A (3GPP NB-IoT), while the second consists of the application of sequential MLAs described in Section IV to the data set from Section III-B (IEEE 802.11n). In the following sections, we discuss the setup and results of the experiments.

\subsection{3GPP NB-IoT}

The total number of instances is 5425, which consists of approximately the same number of fingerprints for each class. Optimization and evaluation of MLAs were performed operating a stratified nested cross-validation procedure \cite{CROSS}. Hyper-parameters optimization was accomplished with a random search technique \cite{RANDOM} and 5-fold inner loop, while the final assessment was done with the outer 10-fold cross-validation. As part of the input, characteristics can be Z-score normalized depending on the MLA requirements. In addition, we measured the model's prediction speed and memory occupied by the prediction.

\begin{figure}[t]
	\centering
	\includegraphics[width=0.9\linewidth]{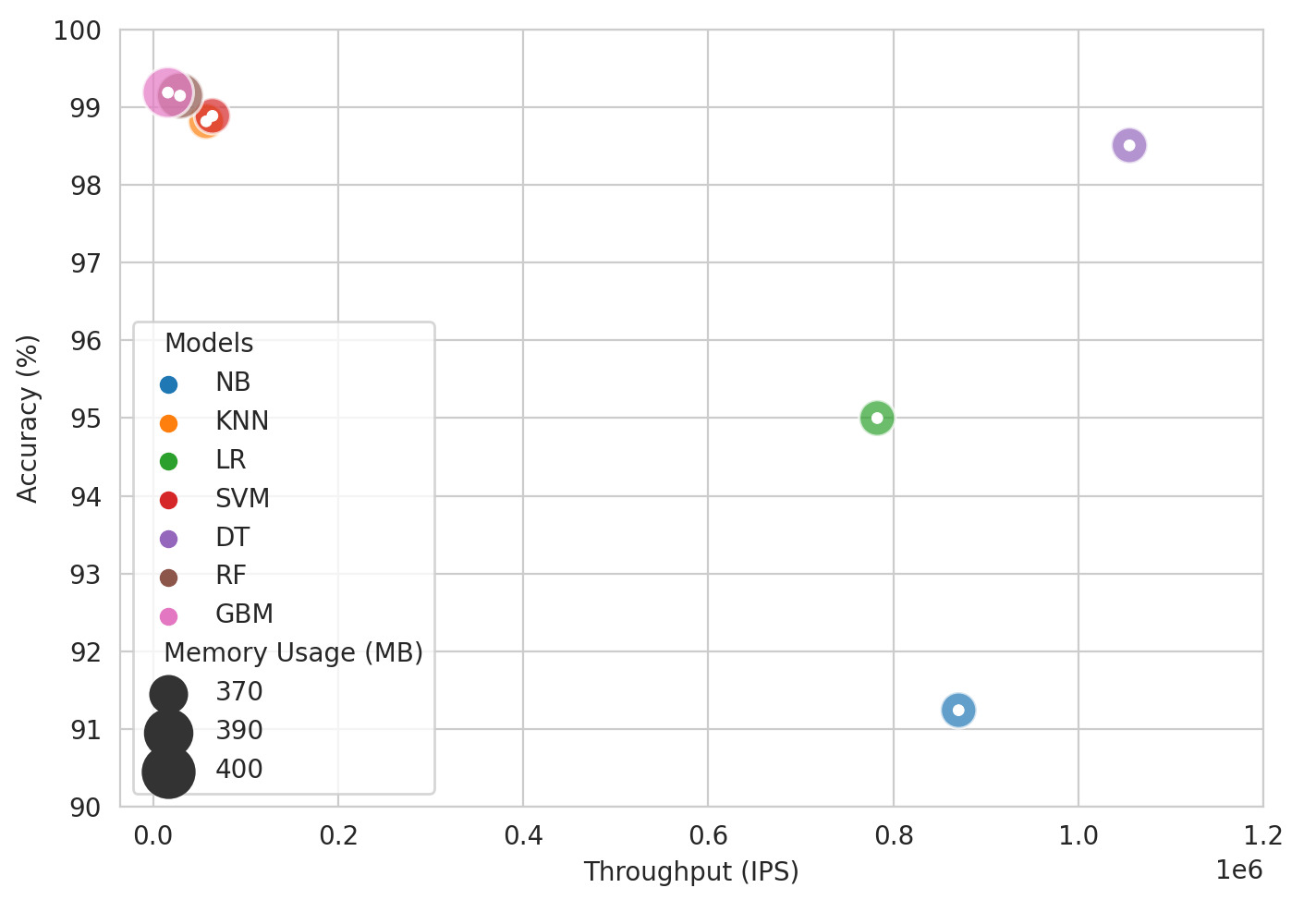}
	\caption{3GPP NB-IoT: Models performance}
	\label{Fig_NB_Results}
\end{figure}

Fig. \ref{Fig_NB_Results} summarizes the performances of models with three principal dimensions. First, the accuracy metric shows the model's effectiveness on the Y-axis, which can be interpreted as the probability of identifying a device. In contrast, the model's efficiency is displayed on the X-axis utilizing the throughput metric, whose unit is instances per second (IPS). The size of the circles indicates the third dimension, which reveals the memory capacity required to load the model and perform a prediction. Detailed results with a 99\% confidence interval are in Table \ref{table_results_NB}.

Only ensemble models achieved an accuracy of over 99\%. The GBM model has the best score of 99.19\%, while RF is in second place with 99.15\% accuracy, but it is more efficient than GBM, even without a parallelization. The most effective standalone models are SVM with 98.89\% and KNN with 98.82\%, which are similar in prediction speed. In terms of throughput, the fastest model is DT, with over 1 million predictions per second, while the LR and NB are the most efficient in the context of memory usage.

\begin{table*}[ht!]
\centering
\caption{3GPP NB-IoT: Models performance}
\begin{tabular}{c c c c c}
\toprule
Model & Accuracy (\%) & F-measure (\%) & Throughput (IPS) & Memory (MB) \\ 
\midrule
NB & {91.24 $\pm$ 0.64} & {87.01 $\pm$ 0.97} & \num{8.70e5} $\pm$ \num{6.38e4} & {363.40 $\pm$ 2.18} \\
LR & {95.00 $\pm$ 0.79} & {91.22 $\pm$ 1.24} & \num{7.82e5} $\pm$ \num{1.09e5} & \textbf{363.09 $\pm$ 1.51} \\
DT & {98.51 $\pm$ 0.42} & {97.63 $\pm$ 0.72} & {\num[math-rm=\mathbf]{1.05e6} $\pm$ \num[math-rm=\mathbf]{6.54e4}} & {365.30 $\pm$ 2.80} \\
KNN & {98.82 $\pm$ 0.40} & {98.16 $\pm$ 0.63} & \num{5.73e4} $\pm$ \num{6.98e3} & {367.74 $\pm$ 2.61} \\
SVM & {98.89 $\pm$ 0.53} & {98.39 $\pm$ 0.81} & \num{6.40e4} $\pm$ \num{3.43e3} & {363.78 $\pm$ 1.95} \\
RF & {99.15 $\pm$ 0.27} & {98.73 $\pm$ 0.43} & \num{2.90e4} $\pm$ \num{1.24e3} & {387.69 $\pm$ 0.46} \\
GBM & \textbf{99.19 $\pm$ 0.27} & \textbf{98.77 $\pm$ 0.39} & \num{1.59e4} $\pm$ \num{6.20e2} & {395.61 $\pm$ 0.57} \\
\bottomrule
\end{tabular}
\label{table_results_NB}
\end{table*}

Based on the mean decrease in impurity, DT has three main predictors: Signal power, Total power, and Energy last packet, with scores of 0.42, 0.40, and 0.14, respectively. Furthermore, the same order is also achieved with the metric based on game theory, average impact on model output magnitude (SHAP values) \cite{SHAP}, with the following scores: 0.83, 0.75, and 0.45.

\subsection{IEEE 802.11n}

The number of generated fingerprints is 24722. The balanced data set was divided with a 50-50 ratio into a training and testing set with the stratified hold-out strategy for all classical MLAs. During their optimization, the 5-fold cross-validation method was operated for evaluation. In contrast, due to the computational complexity in optimizing the deep learning algorithms (DLAs), the data set was split into training (45\%), validation (5\%), and testing (50\%) subset. Such setup indicates that all MLAs were tested on the identical test set. Thus, the final assessment was performed on a held-out test set. In addition, a stratified 9-iteration repeated hold-out procedure was conducted. At each iteration, samples were taken in a predetermined ratio with the corresponding random seed, and models with optimal hyper-parameters were trained and tested with an appropriate sample. In the context of throughput, the model's prediction is measured without utilizing the parallelization of CPU or GPU.

\begin{table*}[ht!]
\centering
\caption{IEEE 802.11n: Models performance}
\begin{tabular}{c c c c c}
\toprule
Model & Accuracy (\%) & F-measure (\%) & Throughput (IPS) & Memory (MB) \\ 
\midrule
WEASEL & {97.27 $\pm$ 0.18} & {97.28 $\pm$ 0.18} & \textbf{481.82 $\pm$ 18.23} & {851.51 $\pm$ 10.97} \\
ROCKET & {99.17 $\pm$ 0.08} & {99.17 $\pm$ 0.08} & {445.43 $\pm$ 10.79} & \textbf{489.98 $\pm$ 2.04} \\
RNN & {99.17 $\pm$ 0.15} & {99.17 $\pm$ 0.15} & {422.30 $\pm$ 8.17} & {4042.51 $\pm$ 3.01} \\
RESNEXT & {99.35 $\pm$ 0.20} & {99.35 $\pm$ 0.21} & {83.05 $\pm$ 0.97} & {4067.66 $\pm$ 15.17} \\
LSTM & {99.42 $\pm$ 0.25} & {99.42 $\pm$ 0.25} & {193.44 $\pm$ 3.91} & {4167.18 $\pm$ 39.65} \\
SKNET & \textbf{99.47 $\pm$ 0.10} & \textbf{99.47 $\pm$ 0.10} & {43.74 $\pm$ 0.68} & {4067.71 $\pm$ 16.27} \\
\bottomrule
\end{tabular}
\label{table_results_ESP}
\end{table*}

All DLAs were optimized with the RMSProp algorithm \cite{RMSPROP} and a batch size of 32. In order to reinforce numerical stability, all attributes were standardized. The learning process begins with the optimized initial value of a learning rate (LR), which is adapted using an LR factor and a plateau-based scheduler. In order to avoid overfitting, the duration of the learning process is determined with early stopping methods if there is no improvement in the model's accuracy at the validation set after 30 consecutive epochs. The regularization techniques dropout and weight decay were used during training.

\begin{figure}[t]
	\centering
	\includegraphics[width=0.9\linewidth]{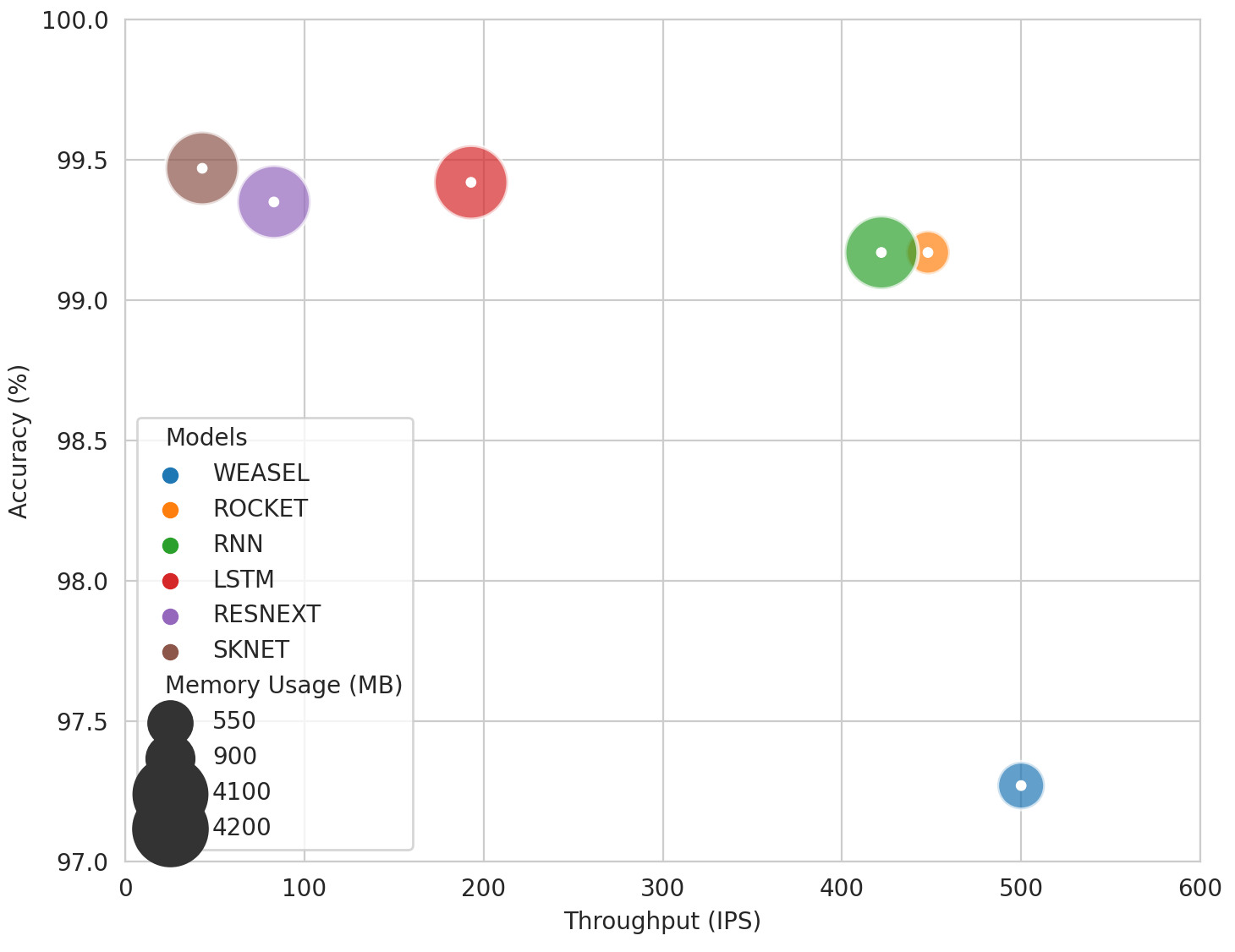}
	\caption{IEEE 802.11n: Models performance}
	\label{Fig_ESP_Results}
\end{figure}

The results show that DL models outperform classical ML models with non-plain convolutional and recurrent architectures. The SKNET model achieved the best result with an accuracy of 99.47\%, which LSTM follows with 99.42\%, and then RESNEXT with 99.35\%. On the other hand, the most efficient model in the DL category is the plain RNN. The best non-DL classifier is ROCKET, with 99.17\% accuracy, which is also the most lightweight model that requires less than 500 MBs of memory. The fastest model is WEASEL, with nearly 500 predictions per second. Table \ref{table_results_ESP} contains detailed results with a 99\% confidence interval, while their visualization is in Fig. \ref{Fig_ESP_Results}.

\section{Conclusion}

In this paper, we presented a detailed study of machine learning algorithms for WF-based device identification. We have implemented a complete end-to-end system based on two major IoT technologies: 3GPP NB-IoT (cellular IoT) and IEEE 802.11n (Wi-Fi IoT). The system is implemented in lab environment with static IoT devices and APs. Besides showing that in a static industrial IoT scenario, device identification using WF is highly efficient, we discussed a large number MLAs for both scenarios and compare their main properties (accuracy, F-measure, throughput and memory footprint). This paper presents a preliminary set of results of a comprehensive study of WF-based device identification that will be done as part of H2020 COLLABS project. The deployment, integration and evaluation of the system in the real-world industrial IoT environment is our ongoing step towards this goal.



\end{document}